\documentclass[conference]{IEEEtran}
\IEEEoverridecommandlockouts
\usepackage{stfloats}
\usepackage[square,numbers]{natbib}
\usepackage{amsmath,amssymb,amsfonts}
\usepackage{algorithmic}
\usepackage{multicol,lipsum}
\usepackage{balance}
\usepackage{caption}
\newcommand{\source}[1]{\caption*{\footnotesize{Source: {#1}} }}
\usepackage{graphicx}
\usepackage{textcomp}
\usepackage{multirow}
\usepackage{xcolor}
\def\BibTeX{{\rm B\kern-.05em{\sc i\kern-.025em b}\kern-.08em
    T\kern-.1667em\lower.7ex\hbox{E}\kern-.125emX}}

\makeatletter
\def\ps@IEEEtitlepagestyle{%
  \def\@oddfoot{\mycopyrightnotice}%
  \def\@evenfoot{}%
}
\def\mycopyrightnotice{
  {\footnotesize 978-1-7281-6215-7/20/\$31.00 ©2020 IEEE\hfill}
  \gdef\mycopyrightnotice{}
}

\begin{document}

\title{Consumer Demand Modeling During 
COVID-19 Pandemic\\}

\author{\IEEEauthorblockN{Shaz Hoda}
\IEEEauthorblockA{\textit{Emerging Technology} \\
\textit{PwC}\\
New York, USA \\
shaz.hoda@pwc.com}
\and
\IEEEauthorblockN{Amitoj Singh }
\IEEEauthorblockA{\textit{Emerging Technology} \\
\textit{PwC}\\
Mumbai, India \\
amitoj.x.singh@pwc.com}
\and
\IEEEauthorblockN{Anand Rao }
\IEEEauthorblockA{\textit{Emerging Technology} \\
\textit{PwC}\\
Boston, USA \\
anand.s.rao@pwc.com}
\and 
\IEEEauthorblockN{Remzi Ural}
\IEEEauthorblockA{\textit{Consumer Markets} \\
\textit{PwC}\\
Chicago, USA \\
huseyin.r.ural@pwc.com}
\and
\IEEEauthorblockN{Nicholas Hodson}
\IEEEauthorblockA{\textit{Consumer Markets} \\
\textit{PwC}\\
Utah, USA \\
nicholas.hodson@pwc.com}
}

\maketitle

\begin{abstract}
The current pandemic has introduced substantial uncertainty to traditional methods for demand planning. These uncertainties stem from the disease progression, government interventions, economy and consumer behavior. While most of the emerging literature on the pandemic has focused on disease progression, a few have focused on consequent regulations and their impact on individual behavior. The contributions of this paper include a quantitative behavior model of fear of COVID-19, impact of government interventions on consumer behavior, and impact of consumer behavior on consumer choice and hence demand for goods. It brings together multiple models for disease progression, consumer behavior and demand estimation– thus bridging the gap between disease progression and consumer demand. We use panel regression to understand the drivers of demand during the pandemic and Bayesian inference to simplify the regulation landscape that can help build scenarios for resilient demand planning. We illustrate this resilient demand planning model using a specific example of gas retailing. We find that demand is sensitive to fear of COVID-19 – as the number of COVID-19 cases increase over the previous week, the demand for gas decreases - though this dissipates over time. Further, government regulations restrict access to different services, thereby reducing mobility, which in itself reduces demand.\
\end{abstract}


\section{Introduction}
Understanding demand and its drivers in the pre-pandemic era was about understanding underlying demographic characteristics of the customers and short-run price and promotion decisions of the supplier \cite{nijs2001category}. Given that consumer preferences and characteristics change slowly over time and price and promotions have temporary effects on demand, it is simple to use machine learning techniques to understand the drivers of demand and to make short (few weeks), medium (few months) and even in some cases, long run (few years) predictions \cite{bajari2015machine}.
These stable relationships, however, have all broken down during the pandemic. Though traditional factors still determine demand to an extent, new drivers become more important as individuals make decisions that change rapidly with the changing environment. 
This makes demand planning and consequently, planning for production and distribution difficult in consumer markets. Given the lack of certainty in demand, it is important to 1) understand the right demand drivers and 2) to be able to model them correctly so as to get to an accurate understanding and forecast demand under different potential scenarios. Further, these demand forecasts/scenarios need to be available not just at a national level, but also at local levels.\

A substantial body of research on the pandemic has focused on disease progression \cite{mollalo2020gis}\cite{wilder2020modeling}\cite{ji2020prediction}\cite{firth2020} with a few looking at the consequent regulations and their impact on individual behavior \cite{lattanzio2020lifting}\cite{smyth2020lockdowns}\cite{silva2020covid}.  On demand forecasting during the pandemic, researchers have demonstrated that reopening of businesses and the return to normalcy will depend not only on regulations but also on the actions of related businesses, and customers and suppliers as well \cite{balla2020business}. Researchers have also identified drivers of household consumption behavior during the pandemic ranging from demographic characteristics, political orientation, geographic location and to a lesser extent, income \cite{baker2020covid}. Our paper combines these strands of work by examining the interaction of these common demand drivers with regulations and testing these drivers along with others for their significance in predicting demand fluctuations. \

The objective of our paper is to consolidate various factors that are impacting consumer demand during the pandemic, as well as to provide an empirical model to estimate demand at local level. We use a demand modeling example of a gas retail company for this paper. For modeling purposes we use gas demand data for 2019 and 2020 (from a gas retailer operating in USA), the data is available at store level for 1000+ stores in 30+ states in the US. We augment this data with multiple rich data sources and run different models LSTM, SARIMAX and panel regression to arrive at variables that can significantly explain the gas demand in 2020 as percentage of gas demand in 2019 for the same week. We finally choose panel regression for two main reasons - 1) lack of data to properly tune other models for over-fitting and 2) for the high degree of interpretability that panel regression allows. We then simplify the regulation landscape and determine how the simplified regulation landscape influences these drivers to develop scenarios for planning. This approach towards demand modeling for gas, with slight modifications, can be extrapolated to modeling demand for other consumer goods such as restaurants, apparel, groceries etc.

\section{DATA}

As discussed in the previous section there are multiple factors that influence demand and the objective of this paper is to create a well-founded framework for determining demand drivers during the pandemic and testing the same empirically. The drivers are then interplayed with regulations to develop planning scenarios. \

\begin{figure}[htbp]
\caption{Systems view of the pandemic and its impact on society.}
\centerline{\includegraphics{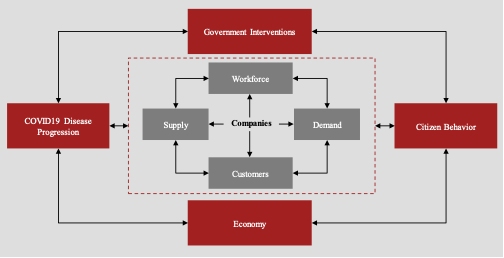}}
\source{Authors' depiction.}
\label{fig1}
\end{figure}

Figure \ref{fig1} is a systems representation of how uncertainties from the disease, government intervention, economy and customer behavior are all intricately linked and cause uncertainty in supply and demand.\

We begin with laying multiple hypotheses for drivers of demand. Through interviews conducted with various experts in the consumer goods industry, a survey of literature and working with varied companies, we break down the potential demand drivers into following categories of data that we collect:\

\subsection{Macro Indicators}

COVID-19 has led to socio-economic effects and behavioral changes because of different policies and regulations adopted by governments across the globe, and because of changes in macro-economic environment of countries – most countries reporting sharp contraction of GDPs for 3rd quarter of FY20 \cite{balla2020business}. These effects have impacted countries and regions, with each regime adopting its own set of regulations . While dips in infections have shown signs of recoveries in countries like Japan and South Korea, a second wave of pandemic and stricter controls have been anticipated in most US states and EU. There has also been a significant variability in the macro-economic response of governments - US and EU countries responded by providing stimulus packages and thereby boosting demand, while developing countries such as India, Mexico, are more fiscally constrained \cite{baker2020covid}. 
These effects are essential for understanding the state-specific impacts of the pandemic on various industries within the USA and therefore, we have included the following set of key variables for the purpose of our analysis. 

\begin{itemize}
    \item GDP growth rate at national/state level for USA from IHS Markit - GDP growth rate is a good indicator of overall demand/income in the economy and can potentially inform demand for individual products
    \item Unemployment rate at the state level for USA from IHS Markit - Unemployment rate directly impacts consumer ability to spend. The job losses have been skewed towards certain sectors – for example, leisure and hospitality being one of the worst-hit sectors. The impact of unemployment was mitigated by the stimulus-package in June, but as its effects wear down, the demand for retail-products also falls significantly \cite{coibion2020cost}. These disparities in job losses affect demand for different goods across different states. 
    \item Personal Consumption Expenditures at state level for USA from IHS Markit - Less disposable income or preference to spend during the uncertain time of a pandemic can adversely impact demand \cite{elgin2020economic}.
    \item COVID-19 Spread from Suspected, Infected, Recovered (SIR) models at USA state level developed by the authors – The SIR model is an epidemiological model indicating the number of infections in a population, based on infection-rates, susceptibility rates and recovery-rates. The spread of infections captures the fear of different individuals in an environment interacting with different factors like strictness of lockdowns, going out and mobility norms \cite{chetty2020did}.
    \item Regulations data at state level for USA from Multistate - Regulations such as closure of bars, restaurants and stores directly affect consumer choice and hence demand for certain types of products.
\end{itemize}

\subsection{Micro Indicators}
It is also important to look at how pandemic has specifically changed consumer habits at an individual level. While the macro effects help explain region specific variations, individual response to the pandemic, has also impacted aggregate consumer demand– mainly via less mobility and more time spent at home. With most white-collared offices moving towards a work from home based approach \cite{coibion2020cost} – significant reduction in mobility levels have been noted across the world. Moreover, reduction in mobility has also, at occasions, resulted in a preference for home-consumable products proved over outside-consumption. For example: we noted that there has been a noticeable increase in sales of salad dressing for consumer goods companies – a 'substitution effect' for visits to restaurants.\
To capture the mobility related effects, we looked at various indicators of mobility for individuals collected through Google Mobility and PlaceIQ. Different indicators of mobility at state/local level can indicate change in behavior of consumer and shifting preferences\
\begin{itemize}
    \item \textbf{Work from Home Index (WFHI) at county level} – We measure work-from-home as a percentage change in time spent at work places compared to the median value in the 5-week period: Jan 3 – Feb 6, 2020. A higher work from home results in fewer trips and less urban mobility \cite{coibion2020cost}. We source WFHI data from Google mobility at a county level and map the WFHI of the county in which the store is located against each store.
    \item \textbf{Stay at Home Index (SAHI) at county level} – Similar to WFHI, SAHI captures the percentage change in time spent at home as compared to the median value in the 5-week period: Jan 3 – Feb 6, 2020. We source SAHI data from Google mobility at a county level and map the SAHI of the county in which the store is located against each store
    \item \textbf{Visits to grocery stores at county level} - Higher visits to grocery stores could be an indicator of consumer preferences shifting towards home cooked meals as compared to restaurants \cite{szymkowiak2020store}\cite{chang2020covid}. In our case, for gas retail, visits to grocery stores were particularly important since the gas stations were generally attached to grocery store chains. We source visits to grocery stores data from PlaceIQ at a county level and map the grocery visits of the county in which the store is located against each store
\end{itemize}

Consumer habits and surveys at county levels - Different indicators of consumer behavior captured through representative surveys
\begin{itemize}
    \item Comfort in going out during pandemic
    \item Activities that consumers miss the most
\end{itemize}

\subsection{Company Data}
We were provided with gas volumes for a company that owns gas-stations across various states in the US. The stores were distributed across 30+ US states, significantly present in rural sites (60\% of total stores). The data was at a weekly level for the 2 years – 2019 and 2020, and we modelled the gas-volumes sold in 2020 as a percentage of sales in 2019 levels for the same period. This was done in order to remove the seasonality-effects, to understand demand fall relative to normal-levels and finally to forecast a recovery compared to 2019 levels. Overall the master data for analysis was created at store week level, with dependent variable as 2020 sales as a proportion of 2019 sales for the same week. For each store the drivers, discussed in previous two sections, were mapped - based on their granularity- to either the county or the state in which the store belonged. 

\section{METHODOLOGY}
With the above data, we wanted to explain and project demand for gas. We followed a three step process to the problem:

\subsection{Testing hypotheses and choosing best fit model}
We used the above set of explanatory variables to understand the impact of the current pandemic on gas-sales. For our analysis, we considered all the macro and micro indicators. However, we leave out regulations from hypotheses testing of initial drivers because all other variables are impacted by state, local and national regulations – we come back to regulations as an overarching variable, once the initial driver model is laid out. For example, the macro variable unemployment is caused due to states mandating closure of certain businesses. Similarly the micro variable stay at home is impacted by closure of business and firms responding to government regulations and guidelines.\

For testing our hypothesis we tried three different machine learning methods:
\begin{itemize}
    \item \textbf{Deep Learning based LSTM method} - Long short-term memory (LSTM) is an artificial recurrent neural network (RNN) architecture \cite{hochreiter1997long}. LSTM networks are well-suited to classifying, processing and making predictions based on time series data.
    \item \textbf{SARIMAX models} - Seasonal Auto Regressive Integrated Moving Averages with Exogenous regressors. Seasonal ARIMAX, is an extension of ARIMA that explicitly supports univariate time series data with a seasonal component and exogenous regressors. It adds three new hyperparameters to specify the autoregression (AR), differencing (I) and moving average (MA) for the seasonal component of the series, as well as additional parameters for the period of the seasonality and exogenous variables.
    \item \textbf{Panel regression model with time series adjustments} - Panel regression tries to overcome the limitations of or ordinary least-squares regressions (OLS) by introducing fixed and random effects variables across individuals (states in our case) and time. As we show in subsequent sections, this model gives us the best fit and interpretability of the model.
\end{itemize}

\subsection{Overlaying impact of regulations}
Once a best fit model is constructed – based on error rates and explainability - we need to develop a model that can inform scenarios on the underlying drivers. For instance, we are most interested in understanding how these variables are interrelated and are determined by overarching factors. For determining the movement of these variables and possible scenarios, we introduce the variable of government regulations. With restrictions passed by governments around mobility of individuals and functioning of businesses, especially in the USA, different macro and micro indicators that we consider are impacted.\

\begin{figure}[htbp]
\caption{Different Regulations and Announcements in one of the US States and associated mobility }
\centerline{\includegraphics[scale=.20]{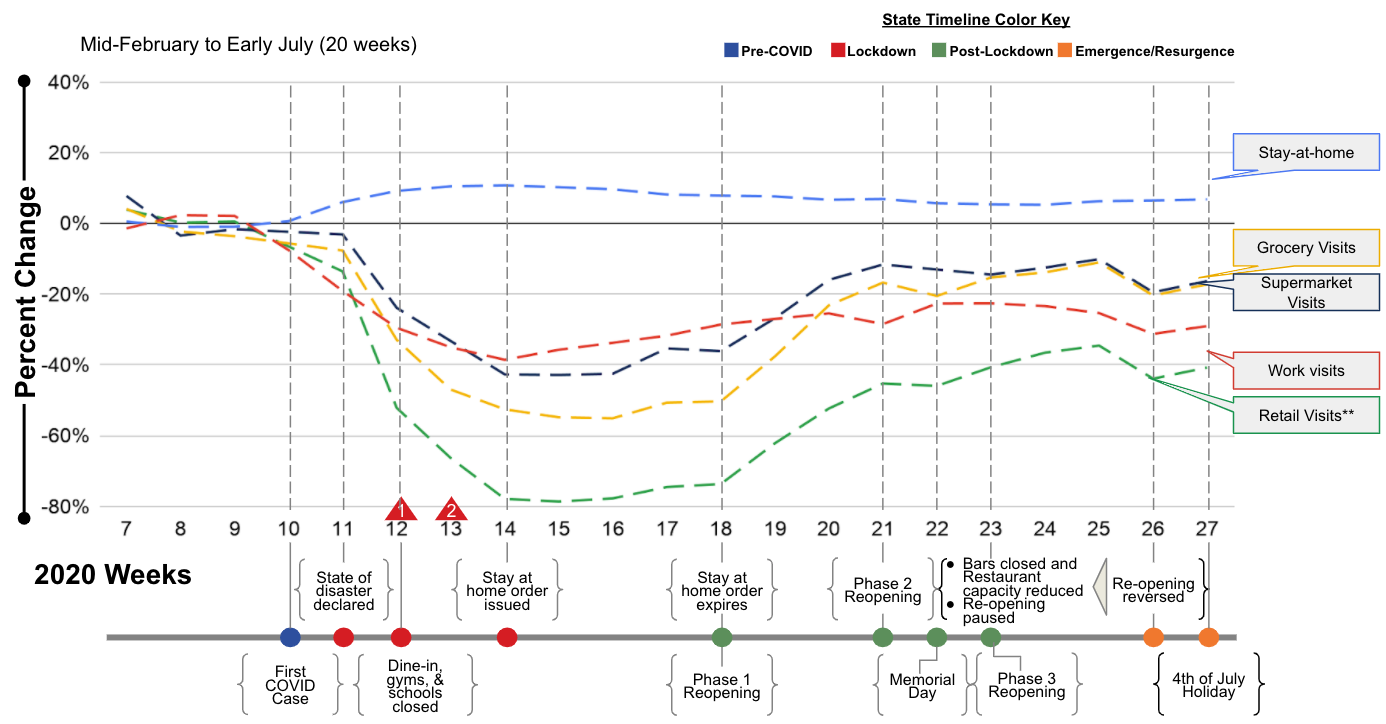}}
\source{Author’s analysis based on Google Mobility and PlaceIQ data}
\label{fig2}
\end{figure}

Figure \ref{fig2} shows how different micro indicators (sourced from Google mobility data) behaved as different regulations and announcements were made in Texas, as an example.\

Given the importance of regulations in determining both macro and micro indicators we procure regulations data in the form of severity of closure/opening of certain businesses, defined with an openness at three different levels. Table 2 (presented in Appendix) shows different types of regulations we capture and how they are classified as Level 1 (mostly open) - Level 4 (mostly closed) based on severity.

\subsection{Building scenarios}
After the models are built and the regulation landscape is set, we use regulations as overarching variables for determining the scenarios and estimating the demand given the regulation level. Through our approach, detailed in a subsequent section, the models learn the dependence between the regulation and our set of micro and macro indicators, taking into account both time-based and location-based effects. Once this is achieved, all the end user needs to do is to specify restrictions for a given time period and location to generate forecasts for the underlying product. This gives the end user the flexibility to plan for both easing and re-imposition of restrictions and hence adopt a resilient strategy for the future. \
The scenario modeling approach involves the following steps: 1) the user chooses expected regulation levels in different US states across time or chooses from pre-built regulatory scenarios (gradual easing/winter lockdown) 2) based on the chosen regulations, we use the mathematical model to arrive at forecasts of drivers 3) we use the forecasted drivers to understand how they can impact demand in the future and create an estimate.

\section{FINDINGS AND RESULTS}
We present in this section of the paper, anonymized results for retail gas demand projections to understand the demand for commonly consumed consumer goods across multiple stores in different states. As discussed, in the previous section we followed the three step approach to understand demand and to build scenarios on it:\
\subsection{Testing hypotheses and choosing best fit model}
\begin{itemize}
    \item \textbf{LSTM} - though promising in terms of model error rates, the models generally have a tendency to quickly overfit unless regularized properly, which can be a tedious task. These models are also black box models which tend to overweight variables that can have a high degree of correlation with the target variable, even though a theoretical foundation may be missing. Hence, we avoid pursuing this method further for demand scenario modeling. We leave LSTM models as a topic to explore further as more data on the pandemic becomes available.
    \item \textbf{SARIMAX model} - This model gives promising results, however, requires careful tuning of seasonality for which a long history of data is required to come up with a reasonable estimation of seasonality and other components. The company had about one and a half years’ worth of data, including the pandemic period. This made it difficult to estimate reasonable models using SARIMAX and have enough data points to validate the model.
    \item \textbf{Panel regression analysis} - We finally went ahead with this approach to estimate demand. These models offer enough flexibility to fine tune variables and to get parameter estimates to help ensure that model coefficients are as expected. The panel time series model helps us capture both location-specific and time-specific attributes, which the other models LSTM (lack of interpretability) and SARIMAX (lack of location attributes) fail to capture. As we were dealing with significant heterogeneity both in terms of locations -1000+ stores spread across 30+ states – and in terms of time based effects in COVID-19 responses and infection – for example: states differed by date of first cases reported and periods of second-wave – controlling for such unobserved factors was important, which a panel model enabled us do in a simpler way. In this section we investigate the results of our panel regression model further. We also overcome the issue of estimation of seasonality by constructing our target variable as change in demand over previous year. The final functional form of our panel regression is as follows:
\end{itemize}

\begin{equation}
\label{eq}
Y_{isct}=\alpha_i + \beta_1 X_{ist}+ \beta_2V_{ic}+ \beta_{3}Z_{ict}+\epsilon_{isct}  
\end{equation}

Where the dependent variable $Y_{isct}$ is the ratio of gas demand in 2020 to 2019 in store \textit{i}, state \textit{s}, county \textit{c}, and time period \textit{t}. The explanatory variables such as unemployment are either at state $X_{ist}$ or such as SAHI at the county ($V_ic$) level.\

As mentioned above, as a precursor to the panel model, we segmented the stores to cluster similar stores together. Taking fall in demand during the first 16 weeks of the pandemic as our target variable, we ran a decision tree on using characteristics underlying the store. These characteristics included multiple demographic information pertaining to the neighborhood in which the stores were located - e.g. average age, household income, ethnicity, rural/urbanness and other such characteristics that are available from the US census. The decision tree model segmented the stores into 3 distinct segments (the client was present primarily in rural and semi-urban markets and not in major cities):

\begin{enumerate}
    \item Primarily rural stores with low population density - These stores saw the least fall in demand.
    \item Primarily urban stores with medium population density with low education prevalence
    \item Primarily urban stores with medium population density with high education prevalence - These stores saw a huge fall in demand, presumably due to higher work from home of urban educated
\end{enumerate}

For each of these segments we created a regression model to understand how different macro and micro economic factors drive demand. There were four factors that came out to be significant at 5\% confidence levels across all three segments:

\begin{itemize}
    \item Fear of COVID-19: COVID-19 infection rate damped exponentially with elapsed time. This variable captures the fear of individuals to freely pursue activities that influence demand of a product. For example, demand for fashionable clothing is low, since people during the first few weeks of pandemic largely stayed at home due to the lack of awareness and riskiness of the virus. This fear, however, dissipated over time. We observed in our data that demand for products goes up over time, even when all else remains the same. That is, if a state was stuck at a certain level of regulation, it would still see an increase in gas demand overtime as people got used to the new normal and found alternate activities. We experimented with multiple functional forms of this variable - linear dampening, log case dampened linearly and exponential damping. The model with exponential dampening of fear gave the best results in terms of meaningful and significant coefficients along with high r-squared. Through our panel approach we test for different dampening effects across different states, however, we don’t observe any significant differences across states in the dampening rate or function. The final fear of COVID-19 variable for each week (t) and each state (i) took the following form: 
    
    \begin{equation*}
        log\dfrac{COVID19 infections_{ist}}{e^{0.12*t}}
    \end{equation*}
    
    The factor of 0.12 for dampening effect came from running multiple iterations to get the best fit model across states with the limited data. With increased data on pandemic, we can better estimate this parameter.
    \item Change in Stay-at-home index (SAHI) vs baseline (\%) of pre-COVID-19 period. This variable captures the impact on demand due to people staying at home. This is different from the fear of COVID-19, as through this variable we try to capture how different regulations affect mobility of people and consequently demand.
    \item Change in supermarket store visits vs year ago (\%). This variable was specifically introduced since the product was specifically linked to sales through physical visits to supermarkets and could not be purchased online.
    \item Change in unemployment rate (\%, absolute from previous years). This variable captures the effect on sales of the product due to macro-economic impact of loss of employment.\
 \end{itemize}   
The results of the regression and estimates of parameters are presented in Table 1. Our target variable was \% change in volume over the previous year and coefficient values indicate meaningful values, confirming our hypothesis.

\begin{table}[hbtp]
\caption{MODEL COEFFICIENTS FOR KEY VARIABLES PREDICTING DEMAND, PANEL DUMMIES EXCLUDED/AVERAGED}
\begin{tabular}{|p{0.35\linewidth}|c|c|c|}
\hline
\multicolumn{1}{|c|}{\multirow{2}{*}{Model variable}}& \multicolumn{3}{c|}{Segments}\\
\cline{2-4} 
\multicolumn{1}{|c|}{} & \multicolumn{1}{c|}{Rural} & \multicolumn{1}{c|}{Urban - educated} & \multicolumn{1}{c|}{Urban - less   educated} \\ 
\hline
Fear of COVID-19: COVID-19 infection rate damped with elapsed   time in weeks & -0.468                     & -0.411                                & -0.058                                       \\ \hline
Change in Stay-at-home index vs baseline (\%)                           & -1.236                     & -1.046                                & -1.597                                       \\ \hline
Change in Grocery store visits vs year ago (\%)                         & 0.292                      & 0.376                                 & 0.178                                        \\ \hline
Change in unemployment rate (\%, absolute)                              & -1.639                     & -0.917                                & -1.151                                       \\ \hline
r-squared                                                               & 0.82                       & 0.90                                  & 0.86                                         \\ \hline
\end{tabular}
\footnotesize{Source: Author’s estimation. All variables significant at 5\% confidence} 
\end{table}

We find that the gas demand across segments decrease by 1 to 1.5 percent with a 1 percent increase in SAHI. In contrast, demand is more sensitive to unemployment changes in less-educated and rural segments as compared to the educated-urban segments. Fear of COVID-19 matters relatively less in the urban, less-educated segment vis-à-vis the others, as perhaps explained by their higher engagement in retail sectors and other essential services.  
Once the estimates are obtained, it is easy to apply these estimates to see how demand moved during the peak of COVID-19 and gradually recovered as these underlying variables changed. We pick as an example one of the major US states, where regulations were significantly eased within eight weeks following the first case of COVID-19, which helped increase overall mobility and consequently demand (Figure \ref{fig2}).

\subsection{Overlaying impact of regulations}
It is important to recognize that most of these drivers themselves are driven by the regulations in place and consumers adapting to those regulations.\
For example, Stay-at-home Index (SAHI), is driven by what is open and closed within the state such as retail stores, restaurants etc. Similarly, visits to supermarkets depend on how restrictive the regulators are in allowing businesses to operate and impose distancing norms. Given the overarching role of regulation in determining these drivers, we develop a mathematical estimate of impact of regulations on drivers.

\begin{figure}[htbp]
\caption{Decline and recovery of Gas Demand for one of the major states during COVID-19 - explained by the demand drivers}
\centerline{\includegraphics[scale=.22]{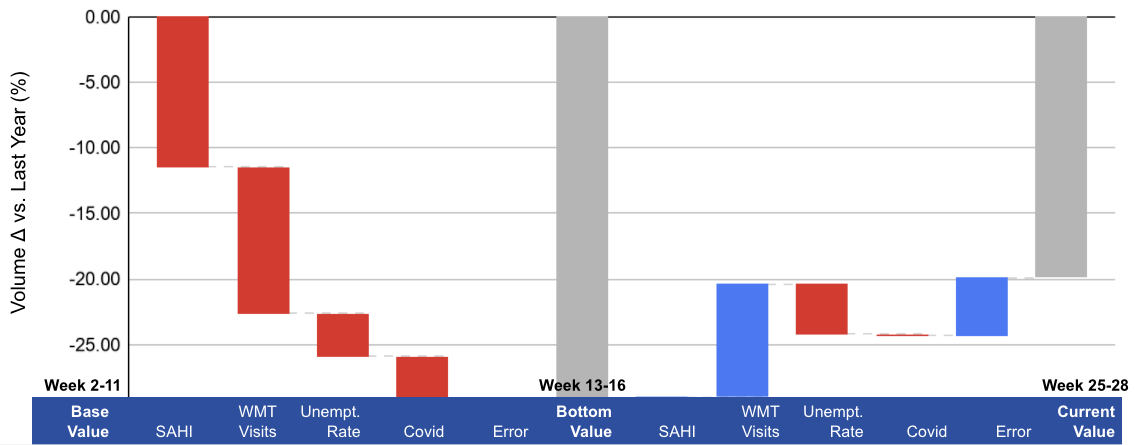}}
\label{fig3}
\end{figure}

To incorporate regulation in our model, we need to classify each state into an overall level of openness. However, as we can see in Table 2 (presented in Appendix), the array of regulations that can be applied in a state and for each of those regulations, the restrictiveness that can be imposed is very complicated. There are a total of 9 different types of regulations and each of these 9 regulations can have 4 levels of imposition. This means that each US state can be in 262,144 ($4^9$) possible regulatory states at any point of time.\
We simplify this complex array of regulations using Bayesian Inferencing to find the hierarchy in which they are applied. For instance, when stay at home order is in place, bars, retail and restaurants are shut as well. Hence regulations are not 

\begin{figure}[htbp]
\caption{Bayesian network of hierarchy of regulations using NO TEARS algorithm}
\centerline{\includegraphics[scale=.65]{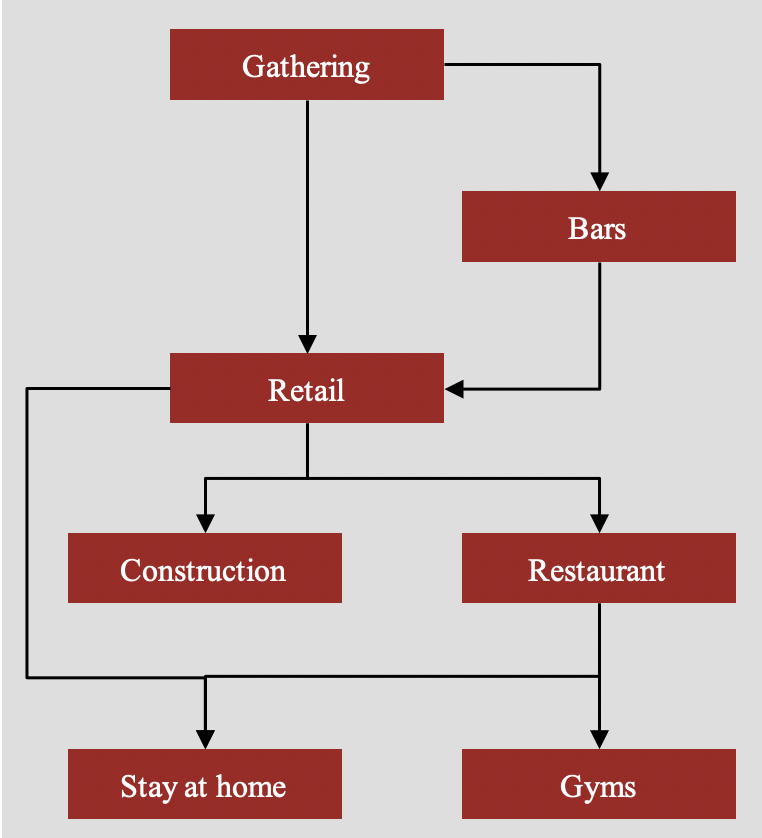}}
\label{fig4}
\end{figure}

randomly applied/removed as a suite of options between the 9 businesses presented in table 2 (presented in Appendix), but more in a hierarchical manner. We use the Non-combinatorial Optimization via Trace Exponential and Augmented lagRangian for Structure learning (NO TEARS) algorithm to learn the structure of regulations \cite{zheng2018dags}. A major challenge of Bayesian Network Structure Learning (BNSL) is enforcing the directed acyclic graph (DAG) constraint, which is combinatorial. While existing approaches rely on local heuristics, NO TEARS introduces a fundamentally different strategy: it formulates the problem as a purely continuous optimization problem over real matrices that avoids combinatorial constraint entirely. The hierarchy of key regulations across businesses is shown through the Bayesian Network obtained in Figure \ref{fig4}.\

\begin{figure}[htbp]
\caption{Classification of different types of regulations across states into levels}
\centerline{\includegraphics[scale=.32]{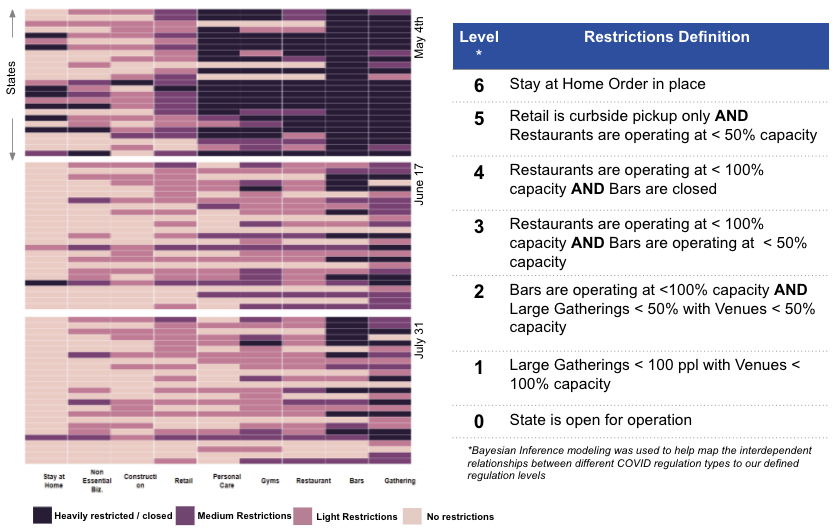}}
\label{fig5}
\end{figure}

Based on the Bayesian hierarchy, the regulations are classified into 6 different levels as presented in Figure \ref{fig5}. These levels are finally used to understand how severity of regulation impacts the key drivers of demand data.

For determining the impact of regulations on drivers, we create a simple mathematical model that depends on regulation level and exponential decay. We estimate the parameter value of the decay through optimization based on available data.\

The two demand drivers which are impacted by regulations are 1) SAHI 2) Grocery Visits. For each of these drivers we fit a mathematical function to determine future behavior based on regulations. The function takes the following form:\

\begin{equation*}
       SAHI_{it}=  \dfrac{SAHI_{i0}*regulationLevelAdjustment}{e^{-k*weeksSinceFirstRegulation}} 
\end{equation*}

Where each \textit{i} represents a different state/county, \textit{t} represents time in weeks and 'regulationLevelAdjustment' represents the regulation level multiplier for each of the 6 different levels of regulation, its value ranging from .1 to 1. We use optimization to estimate the value of k and 'regulationLevelAdjustment' so as to get the best estimate of SAHI (and other demand drivers) based on regulation.

\subsection{Developing Scenarios}
\begin{figure}[hbtph!]
\caption{Density plot of expected demand in different quarters as a percentage of last year demand (x-axis decreasing towards right, range hidden for confidentiality purposes)}
\centerline{\includegraphics{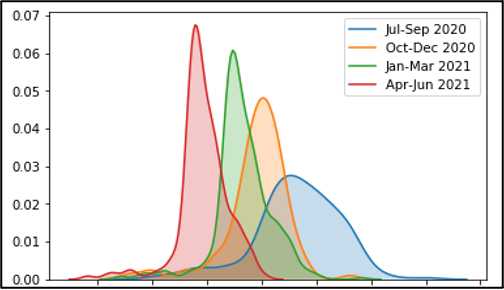}}
\label{fig6}
\end{figure}

Finally using our regression model, overlaid with our regulation-oriented framework for determining demand drivers, we build a scenario modeling tool for the purposes of resiliency planning.
The scenario modeling toolkit allows us to play with different regulatory scenarios and understand the best/worst case forecast along with an overall distribution of forecasts across multiple months, as shown in Figure \ref{fig6}.\
A few interesting insights emerge:
\begin{itemize}
    \item As seen in Figure \ref{fig6}, the overall range and monthly forecast reduces overtime (we hide the x-axis showing demand range to preserve client confidentiality). This is to do with the fact that different states ease regulations overtime and as the pandemic progresses, the initial uncertainty around regulation imposition and compliance becomes more predictable. Further, as we had noted in figure 2, demand gradually recovers overtime as people tend to find new consumption avenues, though the rate of recovery depends on the product being modeled.
\end{itemize}    

\begin{figure}[htbp!]
\caption{Impact of another Lockdown in 2021 (at Level 6) on recovery of gas demand in on one of the US states}
\centerline{\includegraphics{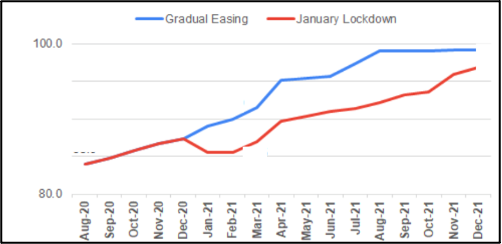}}
\label{fig7}
\end{figure}

\begin{itemize}
    \item The impact of another round of severe lockdown (figure \ref{fig7}), does not impact demand as much as the initial round of severe lockdown observed in early April (demand fell by 25-30\% as compared to previous year). This is again to do with the fact that individuals become resilient in planning and learning. Given the experience in dealing with strict restrictions under a pandemic, consumers find alternate channels and avenues to fulfil demand for goods. 
\end{itemize}

\section{DISCUSSION}
In this paper, we model consumer demand for a gas retailing company under the uncertainties brought on by COVID-19. We provide a quantitative behavioral model of fear around COVID-19 and analyze the impact of regulations on consumer choices and consequently, consumer demand – thus bridging the gap between disease progression and consumer demand. We test the performance of several competing models to identify the drivers of demand and build forecasts based on a range of potential scenarios. \

According to the availability of data, the model is applicable to disaggregated geographic levels. We model demand for a gas retail company, though the same model with slight modifications can be applied (and has been) to modeling demand for other consumer goods, such as restaurants, apparel etc. \

Some limitations and possible extensions to our work would be to address: 1) expanding the scope to include consumer durables (e.g. cars, houses etc.) 2) applying/re-visiting other techniques (LSTM) as more data becomes available 3) use agent based techniques to model long-term behavior changes that can shift demand.
\balance
\begin{figure*}[b]
\centering
    \scalebox{0.5}{\includegraphics{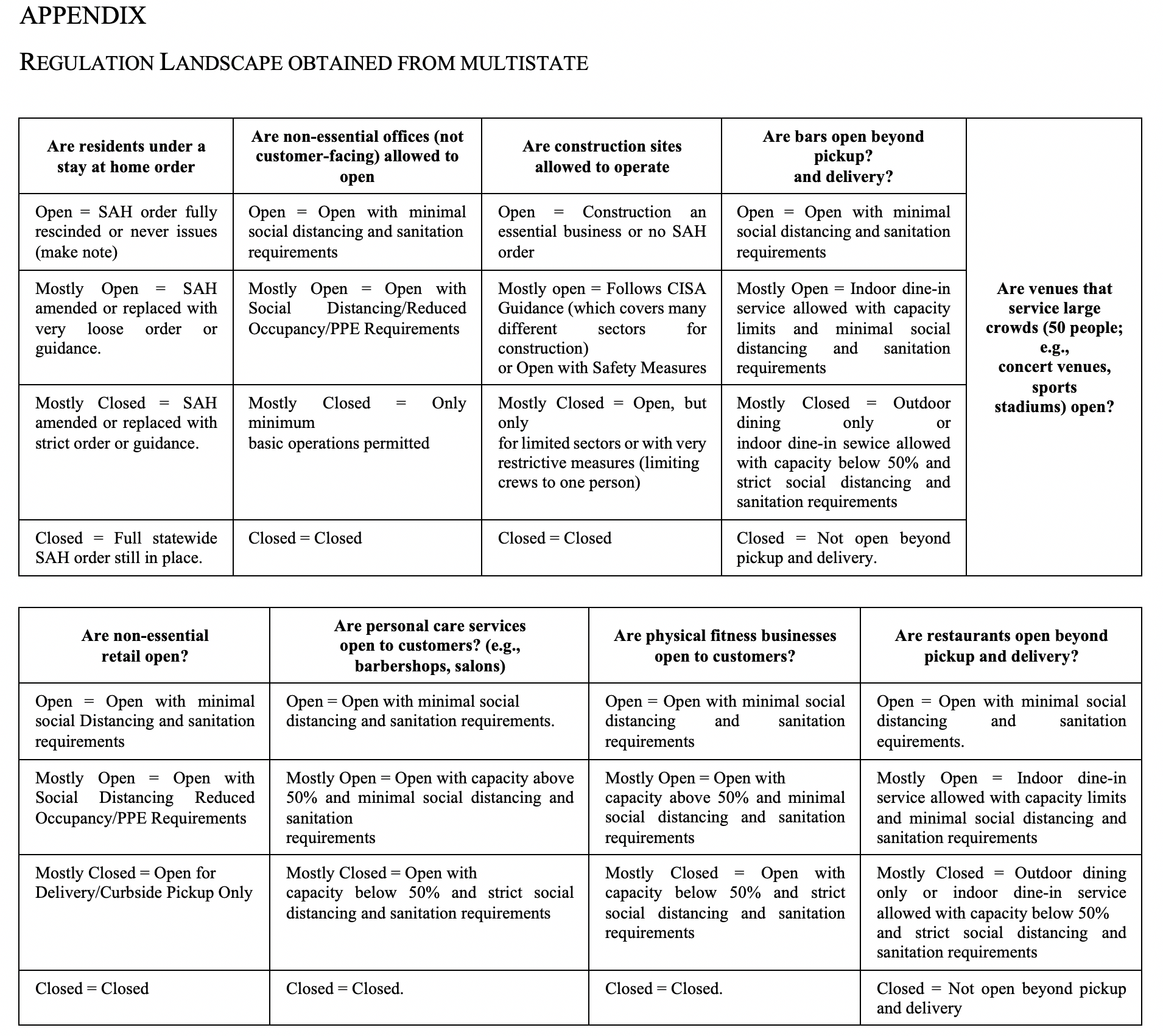}}
\end{figure*}
\bibliography{references.bib}

\end{document}